\documentclass[11pt, a4paper, onecolumn, confidential, copyright, gr]{google}

\usepackage[authoryear, sort&compress, round]{natbib}
\usepackage{microtype}
\usepackage{graphicx}
\usepackage{hyperref}
\usepackage{url}
\usepackage{booktabs}
\usepackage{amsmath}
\usepackage{subcaption}
\usepackage{multirow}     
\usepackage{makecell}     
\usepackage{graphicx}     
\usepackage{wrapfig}

\usepackage{lineno}
\usepackage{hyperref}
\usepackage{listings}
\usepackage{xcolor}
\definecolor{darkblue}{rgb}{0, 0, 0.5}
\hypersetup{colorlinks=true, citecolor=darkblue, linkcolor=darkblue, urlcolor=darkblue}
\definecolor{darkred}{RGB}{150,0,0}
\definecolor{darkgreen}{RGB}{0,150,0}

\usepackage[T1]{fontenc}
\usepackage{enumitem}
\usepackage[capitalize,noabbrev]{cleveref}
\usepackage{listings}
\usepackage{authblk}

\renewcommand\Affilfont{\normalfont\small}
\lstdefinestyle{PythonStyle}{
    language=Python,
    basicstyle=\small\ttfamily,
    commentstyle=\color{gray}\ttfamily,
    keywordstyle=\color{blue}\bfseries,
    stringstyle=\color{orange},
    showstringspaces=false,
    breaklines=true,
    breakatwhitespace=false,
    columns=fullflexible,
    keepspaces=true,
    upquote=true,
    numbers=left,
    numberstyle=\tiny\color{gray},
    numbersep=5pt,
    xleftmargin=2.5em,
    rulecolor=\color{gray}
}

\lstdefinestyle{PromptStyle}{
    language={},
    basicstyle=\small\ttfamily,
    showstringspaces=false,
    breaklines=true,
    breakatwhitespace=false,
    columns=fullflexible,
    keepspaces=true,
    upquote=true,
    xleftmargin=2.5em,
}

\uselogo{} 

\title{Enabling Intrinsic Reasoning over Dense Geospatial Embeddings with DFR-Gemma}

\correspondingauthor{zxuechen@umich.edu, slobodkin@google.com, joydeepp@google.com}

\reportnumber{} 

\renewcommand{\today}{2026-04-07}

\author[12*]{Xuechen Zhang}
\author[1]{Aviv Slobodkin}
\author[1]{Joydeep Paul}
\author[1]{Mandar Sharma}
\author[1]{Samet Oymak}
\author[1]{Shravya Shetty}
\author[1]{Gautam Prasad}
\affil[1]{\thepa{}{ Research}}
\affil[2]{University of Michigan}
\affil[*]{This work was done while the author held a position as a student researcher at Google}

\begin{abstract}
Representation learning for geospatial and spatio-temporal data plays a critical role in enabling general-purpose geospatial intelligence. Recent geospatial foundation models, such as the Population Dynamics Foundation Model (PDFM), encode complex population and mobility dynamics into compact embeddings. However, their integration with Large Language Models (LLMs) remains limited. Existing approaches to LLM integration treat these embeddings as retrieval indices or convert them into textual descriptions for reasoning, introducing redundancy, token inefficiency, and numerical inaccuracies.
We propose Direct Feature Reasoning-Gemma (DFR-Gemma), a novel framework that enables LLMs to reason directly over dense geospatial embeddings. DFR aligns high-dimensional embeddings with the latent space of an LLM via a lightweight projector, allowing embeddings to be injected as semantic tokens alongside natural language instructions. This design eliminates the need for intermediate textual representations and enables intrinsic reasoning over spatial features.
To evaluate this paradigm, we introduce a multi-task geospatial benchmark that pairs embeddings with diverse question–answer tasks, including feature querying, comparison, and semantic description. Experimental results show that DFR allows LLMs to decode latent spatial patterns and perform accurate zero-shot reasoning across tasks, while significantly improving efficiency compared to text-based baselines.
Our results demonstrate that treating embeddings as primary data inputs, provides a more direct, efficient, and scalable approach to multimodal geospatial intelligence.
\end{abstract}

\begin{document}

\maketitle

\section{Introduction}
Geospatial reasoning is a fundamental capability for enabling general-purpose intelligence over real-world environments, with applications spanning urban planning, mobility analysis, disaster response, and location-based services. Such reasoning requires models to process complex, long-context data, integrate multiple modalities (e.g., maps, continuous environmental measurements, remote sensing imagery, time-series population activity, and categorical POI distributions), and generalize across diverse tasks and geographic regions.

Recent advances in geospatial foundation models~\cite{agarwal2024general,zhao2026mora,butsko2025deploying}, such as Population Dynamics Foundation Models~\cite{agarwal2024general} (PDFMs), have made significant progress toward this goal by encoding rich spatio-temporal and population dynamics into dense embeddings. In parallel, Large Language Models (LLMs) have demonstrated strong capabilities in long-context understanding, multi-step reasoning, and cross-domain generalization. These properties make LLMs a promising backbone for geospatial intelligence, where reasoning over heterogeneous signals and adapting to diverse queries are essential.

\begin{figure*}
\centering
\includegraphics[width=0.9\textwidth]{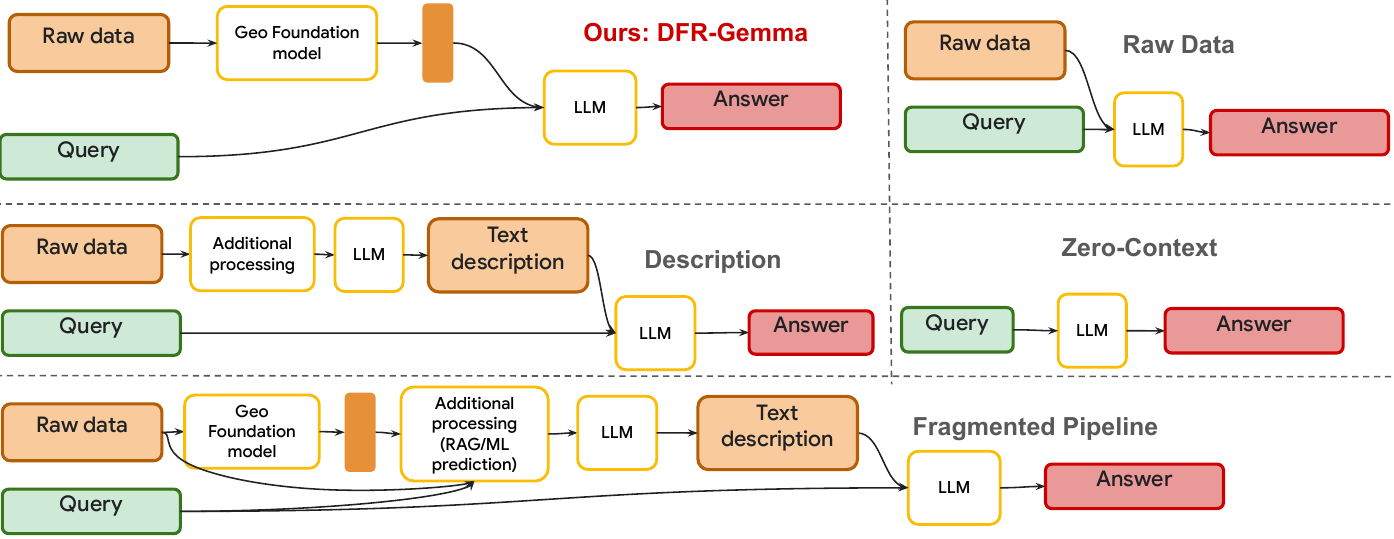}
\caption{Comparison of geospatial reasoning pipelines. DFR (top) enables direct reasoning over embeddings, while prior methods rely on fragmented pipelines or noisy text/raw inputs, resulting in inefficiency and errors.}
\label{fig:fig1}
\end{figure*}

However, a fundamental gap remains: While LLMs are good at linguistic logic, they lack a native mechanism to understand and reason on dense spatial embeddings, creating a critical gap between geo-embedding and natural language reasoning. Existing approaches~\cite{GoogleGeospatial2025, yu2025spatial,zhao2026mora,tucker2024systematic} bridge this gap through indirect and fragmented pipelines. As shown in \Cref{fig:fig1}, existing methods follow a \emph{fragmented pipeline}: embeddings are only used for retrieval (RAG) or processed by intermediate models. These designs are brittle, as errors propagate across stages, and inefficient. To use LLM for reasoning, everything needs to be converted into text, which increases token usage and introduces numerical inaccuracies. Alternatives based on verbose descriptions or raw inputs further degrade performance by overwhelming the context window. These limitations motivate a unified, end-to-end framework for direct reasoning over geospatial embeddings.

To address these limitations, we introduce the Direct Feature Reasoning-Gemma (DFR-Gemma), a framework that enables LLMs to reason directly over geospatial embeddings. Unlike prior approaches, our approach treat the spatial embedding as the primary analytical inputs. Specifically, we project geospatial embeddings into the latent embedding space of a frozen LLM (e.g., Gemma) via a lightweight projector, allowing them to be injected as semantic tokens(soft) alongside natural language instructions. This alignment allows the LLM to reason and interpret over geospatial semantics, such as the relative density of coffee shops versus milk tea shops, or whether population activity suggests a higher frequency of searches for gyms or restaurants, without relying on intermediate textual descriptions or external retrieval. As a result, the model can also directly predict related external features, such as the local unemployment rate. 

To support this research and facilitate future benchmarking, we introduce a multi-task geospatial benchmark that pairs dense embeddings with diverse question–answer tasks, including feature querying, comparison, and semantic description. This setup enables controlled evaluation of cross-modal reasoning\footnote{
We use \emph{multimodal} to denote combining natural language (e.g., queries and instructions) with heterogeneous geospatial signals (e.g., population activity, environmental factors, Point-of-interest(POI) distributions), rather than conventional modalities such as images or audio. These signals are encoded as dense embeddings by a geospatial foundation model and fused with text for joint reasoning.
} and provides a standardized testbed for future research. 

Overall, DFR-Gemma represents a shift from using embeddings as retrieval indices to treating them as primary inputs for reasoning, enabling more direct, efficient, and expressive geospatial intelligence. Our key contributions of this paper are as follows:
\begin{itemize}
\item \textbf{Direct Feature Reasoning Architecture:} We propose a model-agnostic framework that injects geospatial embeddings into the LLM input via a learned projection layer. By aligning embeddings with the LLM latent space and treating them as soft tokens, DFR enables direct reasoning over non-textual data without modifying the backbone, improving token efficiency, robustness, and numerical fidelity.

\item \textbf{Semantic Decoding \& Reasoning:} We show that pre-trained LLMs can decode, verbalize, and reason over dense geospatial embeddings, enabling complex inference without retrieval or intermediate models, with strong generalization across tasks and styles.

\item \textbf{Contextual Compositionality:} We demonstrate that DFR supports dense–sparse hybrid reasoning, enabling seamless integration of geospatial embeddings with large textual contexts for joint reasoning.

\item \textbf{Multi-Task Geospatial Benchmark:} We introduce a dataset bridging geospatial embeddings and language tasks, covering querying, comparison, and description.

\end{itemize}


\section{Related work}
\textbf{Geospatial Representation Learning and Foundation Models}
Representation learning for geospatial and spatio-temporal data has been widely studied in applications such as mobility modeling, urban analytics, and location-based services. Recent geospatial foundation models~\cite{agarwal2024general,zhao2026mora,butsko2025deploying,liu2025selfimputation}, including Population Dynamics Foundation Models (PDFMs)~\cite{agarwal2024general}, learn high-dimensional embeddings that capture rich patterns of human activity, environmental constraints, and socio-economic signals at scale. These embeddings provide a compact and expressive representation of geographic regions, enabling similarity search, clustering, and downstream prediction tasks. However, existing approaches primarily treat such embeddings as static features, without enabling direct reasoning over their internal semantics.

\textbf{Geo-Reasoning via Retrieval-Augmented Generation (RAG):} A common approach to integrate geospatial data with LLMs is RAG. In these frameworks, geospatial embeddings are used as indices to retrieve relevant documents, knowledge graphs, or textual descriptions, which are then provided to the LLM for reasoning \cite{tucker2024systematic,yu2025spatial,Grossetal2025,GoogleGeospatial2025,zhao2026mora}.  This indirect pipeline introduces latency, increases token usage, and is sensitive to retrieval errors. In contrast, we eliminate retrieval and enable direct reasoning over embeddings within the LLM.

\textbf{Multimodal LLMs and Cross-Modal Alignment:} Recent advances in multimodal LLMs have demonstrated the effectiveness of aligning non-textual inputs with language models through shared latent representations. Architectures such as Flamingo, PaLI, and related vision-language models incorporate visual features via cross-attention or learned projection layers \cite{alayrac2022flamingo, beyer2024paligemma, he2022masked,kuomammut,minderer2022simple}. Beyond vision and audio, recent work has explored extending LLMs to structured modalities such as tabular and time-series data using serialization, quantization, or modality-specific encoders \cite{gardner2024large, ansari2024chronos,10.24963/ijcai.2024/921,jiangempowering,bandara2024attention,zhou2023one,jintime}. However, these advances focus on generic modalities and do not address geospatial reasoning. In this domain, there is a lack of datasets that define language-based reasoning tasks. It's unclear whether LLMs can interpret over geospatial embeddings. We address this gap by introducing a benchmark and reasoning framework.

\textbf{Continuous and Hybrid Token Representations for LLM Reasoning:}
Recent work explores continuous or hybrid token representations that mix latent and textual signals~\cite{su2025token,gozeten2025continuous,hao2024training,liu2025hamburger,shen2025codi}. These approaches show that a single token can encode rich semantic information beyond discrete symbols. Such tokens can capture complex semantic structures, intermediate reasoning states, or compressed feature representations. Building on this insight, we treat structured geospatial embeddings as tokens in the LLM latent space, enabling direct reasoning over features rather than relying on textual serialization or retrieval pipelines.

\section{The Direct Feature Reasoning-Gemma (DFR-Gemma) Paradigm} \label{sec:paradigm}
\begin{figure*}
\centering
\includegraphics[width=1\textwidth]{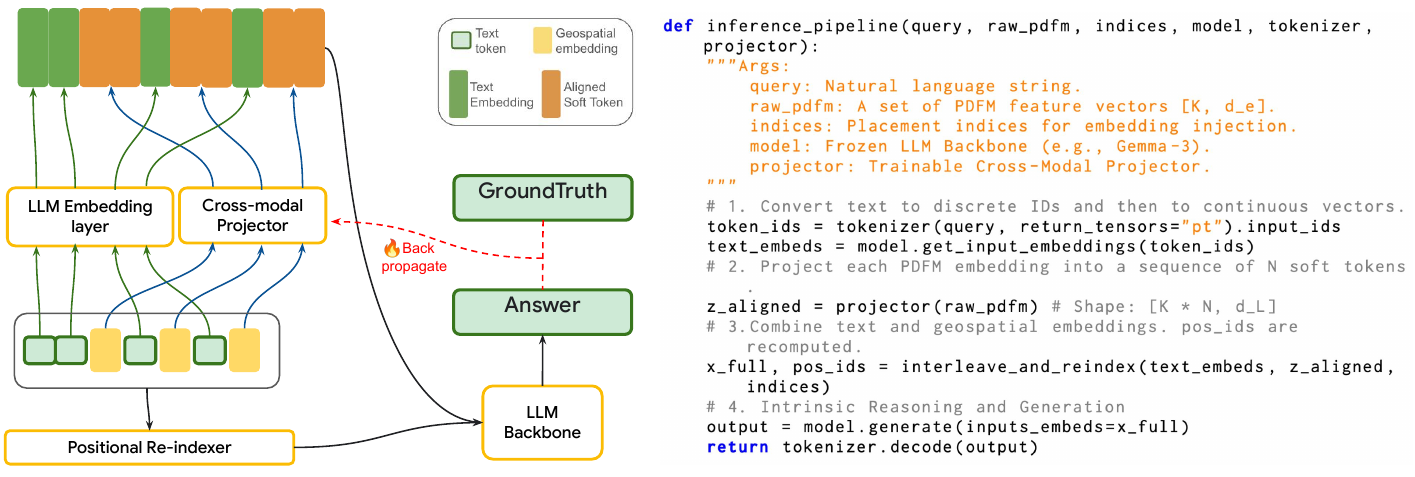}
\caption{The architecture (left) and corresponding pseudocode (right) illustrate the end-to-end integration of geospatial features into a frozen LLM. Detaied explaination are in \Cref{sec:paradigm} and \Cref{sec:method}. In DFR-Gemma, the Projector maps dense each PDFM embedding directly into $N$ Aligned Soft Tokens. We shown $N=2$ as an example in the figure. To retain pre-trained linguistic intelligence, the LLM Backbone remains frozen while only the Cross-Modal Projector is optimized to align the geospatial manifold with the LLM's latent space. During inference, the Positional Re-indexer recomputes sequence IDs for the interleaved text and soft tokens. This ensures the attention mechanism correctly interprets spatial-textual relationships, enabling the model to decode latent patterns and perform intrinsic, zero-shot reasoning validated against ground truth.}
\label{fig:model_framework}
\end{figure*}

In this section, we formalize our approach for enabling LLMs to reason over geospatial information without relying on verbose textual descriptions, which are often inefficient and lossy for representing structured geo data. We encode geospatial signals into embeddings and fuse them with text inputs (queries and instructions), enabling joint reasoning.  The framework and Pseudocode is shown in \Cref{fig:model_framework}.
\textbf{Mathematical Formulation:} Let $\mathcal{R} = \{r_i\}_{i=1}^k$ denote a set of geographic regions with associated raw signals (e.g., population activity, environmental factors, and POI distributions). We first encode each region into a dense embedding using a geospatial foundation model $f_{\text{geo}}: \mathcal{R} \rightarrow \mathbb{R}^{d_e}$, getting $\mathcal{E} = \{e_i\}_{i=1}^k$ where $e_i = f_{\text{geo}}(r_i) \in \mathbb{R}^{d_e}$, each $e_i \in \mathbb{R}^{d_e}$ encodes regional signals. To bridge the modality gap between geospatial embedding and text, we define a Cross-Modal Projector $\phi: \mathbb{R}^{d_e} \to \mathbb{R}^{N \times d_{llm}}$ that transforms each individual regional embedding into a sequence of $N$ Aligned Soft Tokens:
$$Z_i = \phi(e_i) = \{z_{i,1}, z_{i,2}, \dots, z_{i,N}\}$$
where each $z_{i,n} \in \mathbb{R}^{d_{llm}}$ resides in the LLM's latent space. The final unified input to the LLM Backbone is a \textbf{mixed-modality sequence} $X = [x_1, x_2, \dots, x_T]$, where each element $x_t$ is defined as:
$$x_t = 
\begin{cases} 
\text{Emb}(w_t) & \text{if input is a discrete text token } w_t \in \mathcal{V} \\
z_{i,n}\in Z_i & \text{if input is the }\text{ geospatial
embedding } e_i
\end{cases}$$
Here, $\mathcal{V}$ represents the discrete "hard" tokens of the LLM vocabulary. $\text{Emb}(\cdot)$ denotes the frozen LLM Embedding Layer mapping discrete vocabulary tokens $w_t$ to the latent space. This formulation explicitly shows how DFR expands compact geospatial data into a multi-token representation.

\textbf{Technical Challenges:} This paradigm is specifically designed for tasks where natural language and latent features are tightly coupled, such as:\textit{``Given the feature vector \texttt{<emb>$e_0$</emb>} for Region A, which of the following: (1) \texttt{<emb>$e_1$</emb>}, or (2) \texttt{<emb>$e_2$</emb>} represents a similar socio-economic profile?''} The DFR setting introduces two primary hurdles beyond existing work:
\begin{itemize}[nosep,leftmargin=*]
\item \textbf{Cross-Modal Alignment:} The model must bridge the distributional shift between the continuous, structured space of the geospatial foundation model and the discrete, semantic space of the LLM.
\item \textbf{Latent Semantic Decoding:} The model must perform logical operations over interleaved tokens to extract and compare information "locked" within the embeddings (e.g., $e_0, e_1, e_2$). By successfully performing \textbf{comparative latent decoding}, the model generates grounded linguistic responses without relying on external retrieval or textual descriptors.
\vspace{-5pt}
\end{itemize}

\section{Preliminary: Population Dynamics Foundation Model (PDFM)}
The Population Dynamics Foundation Model (PDFM)~\cite{agarwal2024general} serves as the core geospatial feature extractor $f_{\text{geo}}(\cdot)$ for our DFR-Gemma. We utilize PDFM as it represents the state-of-the-art (SOTA) in geospatial foundation models, moving beyond task-specific applications like navigation or satellite imagery to learn universal representations of geographic space. We illustrate the PDFM framework, including example inputs and detailed explanations, in \Cref{fig:pdfm}.

\textbf{GNN-based Representation:} 
The PDFM utilizes a Graph Neural Network (GNN) architecture to learn the geospatial representations. The model constructs a large-scale, heterogeneous graph in which nodes represent geographic units (such as U.S. postal codes and counties) and edges represent spatial proximity or functional similarity. Through self-supervised training, the GNN compresses these diverse signals into a fixed-dimensional latent vector $e \in \mathbb{R}^{330}$. These embeddings encapsulate a "distilled" understanding of a region’s population dynamics, providing a more robust and meaningful input for DFR-Gemma than raw numerical coordinates or sparse categorical data.

\textbf{Encoded Features:}
The primary advantage of PDFM is its ability to distill heterogeneous "maps data" into a unified representation. These embeddings encode a diverse set of real-world features:
\begin{itemize}
\item \textbf{Population-centric data:} Aggregated search trends and activity levels (e.g., location busyness) capturing regional interests and human dynamics.
\item \textbf{Environmental data:} Weather conditions and air quality measurements influencing local behavior.
\item \textbf{Local characteristics:} Point-of-interest categories describing available amenities, services, and infrastructure. (e.g., dense clusters of restaurants, cafes)
\vspace{-5pt}
\end{itemize}
This richness allows DFR-Gemma to answer complex, human-centric questions, such as "Are there more coffee shops or milk tea shops in this area?" or "Which of these two regions is currently busier?", directly from the latent vector. By using PDFM, we can leverage paired raw data to generate ground-truth labels for these tasks, ensuring that the model’s linguistic output is grounded in verifiable truths.

\section{Methodology: DFR-Gemma} \label{sec:method}
While the DFR paradigm defined in Section 4 provides a general framework for feature-based reasoning, this section details our specific implementation. The main framework is shown in \Cref{fig:model_framework}.
\subsection{Model Architecture and Projector Alignment}
We implement the mapping function $\phi: \mathbb{R}^{d_e} \to \mathbb{R}^{N \times d_{llm}}$ using a multi-layer perceptron (MLP) with a terminal expansion layer. To bridge the modality gap, we project each PDFM embedding into $N$ soft tokens. We project each PDFM embeddings $e \in \mathbb{R}^{d_e}$ into a sequence of $N$ "soft tokens" $Z_i = \{z_{i,1}, \dots, z_{i,N}\}$. The projector utilizes a Multi-Layer Perceptron (MLP) architecture with a GELU activation function and a terminal expansion layer. Formally, for an input embedding $e_i$, the projected sequence is computed as:
$$Z_i = \text{Reshape}\left( W_2 \cdot \text{GELU}(W_1 e_i + b_1) + b_2, (N, d_{llm}) \right)$$
where $W_1 \in \mathbb{R}^{d_{mid} \times d_e}$ and $W_2 \in \mathbb{R}^{(N \cdot d_{llm}) \times d_{mid}}$ denote the weight matrices, and $d_L$ is the latent dimension of the LLM Backbone. 
The use of $N \geq 1$ tokens is a deliberate design choice in our method to solve two specific problems:
\begin{itemize}
\item \textbf{Information Density:} A single LLM token lacks the capacity to represent the multi-modal richness (POI, busyness, search trends) of a PDFM embedding.
\item \textbf{Multi-Task Adaptability:} 
Since we utilize a single universal projector for our entire multi-task benchmark, $N$ tokens provide the model with increased "latent bandwidth," allowing the transformer's attention mechanism to selectively extract the task-relevant features needed for a specific query, which would be bottlenecked by a single-token representation. 
\end{itemize}
In our experiments, we evaluate how increasing $N$ improves performance across varied reasoning categories.
\subsection{Mixed-Modality Sequence Construction}
The core of DFR-Gemma is the construction of the interleaved input sequence $X$. We use special placeholder tokens \texttt{<emb>} to mark insertion points for geospatial features within the text. At inference, DFR-Gemma acts as a plug-and-play extension to standard transformers by bypassing the tokenizer and injecting continuous spatial embeddings directly into the model’s first layer.

\textbf{Interleaving Mechanism:} The input text is first tokenized into embeddings. When a placeholder is encountered, the corresponding PDFM embedding is projected via $\phi$ into $N$ continuous vectors, which are inserted into the hidden sequence.

\textbf{Positional Encoding:} Inserting $N$ tokens shifts subsequent positions, so we apply dynamic re-indexing. Each soft token $z_{i,n}$ is assigned a unique positional ID, ensuring correct relative positioning between text and geospatial features in self-attention.
\vspace{-5pt}
\subsection{Training Objective}
\vspace{-5pt}
The framework is trained via supervised fine-tuning on our multi-task geospatial benchmark using cross-entropy loss 
$\mathcal{L}=-\sum_{j=1}^{M}\log P(y_j\mid y_{<j},X;\theta_{\phi})$, where $\theta_{\phi}$ denotes the projector parameters. This enables the projector to decode latent spatial signals in $e$ and perform reasoning. An ablation with supervised contrastive loss (\Cref{app:Contrastive}) shows that cross-entropy alone is sufficient.
\subsection{Data Collection and QA Generation}
To facilitate cross-modal learning, we constructed a comprehensive multi-task geospatial dataset pairing PDFM embeddings with verifiable question and ground-truth answer pairs. To generate the data, we extract raw features, generate corresponding QA pairs, and apply semantic augmentation to diversify linguistic expressions and mitigate template overfitting. The detailed pipeline with prompt and generated examples are shown in \Cref{app:data}.

\textbf{Included tasks:} The DFR architecture is trained and evaluated through a suite of geospatial reasoning tasks designed to teach and test the model's ability to synthesize, compare, and infer knowledge from high-dimensional embeddings. Beyond simple tasks such as classification, DFR-Gemma must bridge the gap between latent space and linguistic logic, performing multi-step cognitive operations to interpret complex spatial contexts. There are three types of queries:
    \begin{itemize}[nosep,leftmargin=*]
    \item \textbf{Single-Embedding Queries:} These tasks evaluate the model’s ability to decode and reason over a single PDFM embedding. Example: \texttt{"As shown in \texttt{<emb>}$e_0$\texttt{</emb>}, there are more coffee shops or milk tea shops?" $\rightarrow$ "Lower"}. The model must perform comparative reasoning by extracting the relevant signal and determining which one is higher.

    \item \textbf{Feature Description:} This task resembles geospatial captioning, requiring the model to translate embeddings into a coherent narrative. It involves semantic reasoning to prioritize important features and produce concise summaries.
    \item \textbf{Multi-Embedding Queries:} This task evaluates the model’s ability to jointly reason over multiple PDFM embeddings (e.g., across regions or time). It requires relational reasoning, as the model must compare and infer across embeddings rather than decode them independently. Example: "\texttt{
Given the feature vector \texttt{<emb>}$e_0$\texttt{</emb>}, which of 
\texttt{<emb>}$e_1$\texttt{</emb>} \{Santa Clara\},
\texttt{<emb>}$e_2$\texttt{</emb>} \{San Mateo\},
\texttt{<emb>}$e_3$\texttt{</emb>} \{Alameda\}, or 
\texttt{<emb>}$e_4$\texttt{</emb>} \{New York\} 
is most similar in terms of weather?} The model must align embeddings with regions, extract the relevant feature (weather), and perform comparative analysis to identify the closest match.
\vspace{-5pt}
\end{itemize}

We generate diverse queries. More examples and detailed question type discussion are given in \Cref{app:data_task}.

\section{Experiments}

\begin{table*}[t]
\centering
\resizebox{1\textwidth}{!}{%
\begin{tabular}{ll cc cc cc} 
\toprule
\multirow{2}{*}{\textbf{Type}}& \multirow{2}{*}{\textbf{Method}} & 
\multicolumn{2}{c}{\textbf{Single-Embed Queries}} & 
\multicolumn{2}{c}{\textbf{Feature Description}} & 
\multicolumn{2}{c}{\textbf{Multi-Embed Queries}} \\
\cmidrule(lr){3-4} \cmidrule(lr){5-6} \cmidrule(lr){7-8} 
& & \textbf{Gemma} & \textbf{Qwen} & \textbf{Gemma} & \textbf{Qwen} & \textbf{Gemma} & \textbf{Qwen}  \\
\midrule
\multirow{3}{*}{No Training} & Zero Context & 0.67 & 0.54 & 173.59 & 254.30 & 0.21 & 0.24\\
& Unprocessed Raw Input & 0.63 & 0.61 & 284.81 & 298.03 & 0.18 & 0.30  \\
& Raw Data Description & 0.70 & 0.76 & 152.45 & 351.87 & 0.46 & 0.53  \\
\midrule
\multirow{2}{*}{No LLM} & MLP & \multicolumn{2}{c}{0.76} & \multicolumn{2}{c}{/} & \multicolumn{2}{c}{0.37} \\
& LightGBM & \multicolumn{2}{c}{0.81} & \multicolumn{2}{c}{/} & \multicolumn{2}{c}{0.39}   \\
\midrule
Ours & DFR-Gemma, N = 4 & 0.79 & 0.76 & 21.03 & 16.84 & 0.72 & 0.58  \\
\bottomrule
\end{tabular}
}
\caption{Performance Comparison on Geospatial Reasoning Tasks. Performance of description is measured by perplexity(ppl). Others are measured by Accuracy (\%). DFR-Gemma consistently outperforms the baselines, particularly in complex multi-modal tasks, demonstrating the effectiveness of reasoning directly over latent geospatial embeddings compared to textual serialization.}
\label{tab:qa_results}
\end{table*}

\begin{figure*}
  \begin{subfigure}[b]{0.32\linewidth}
    \centering
    \includegraphics[width=1\linewidth]{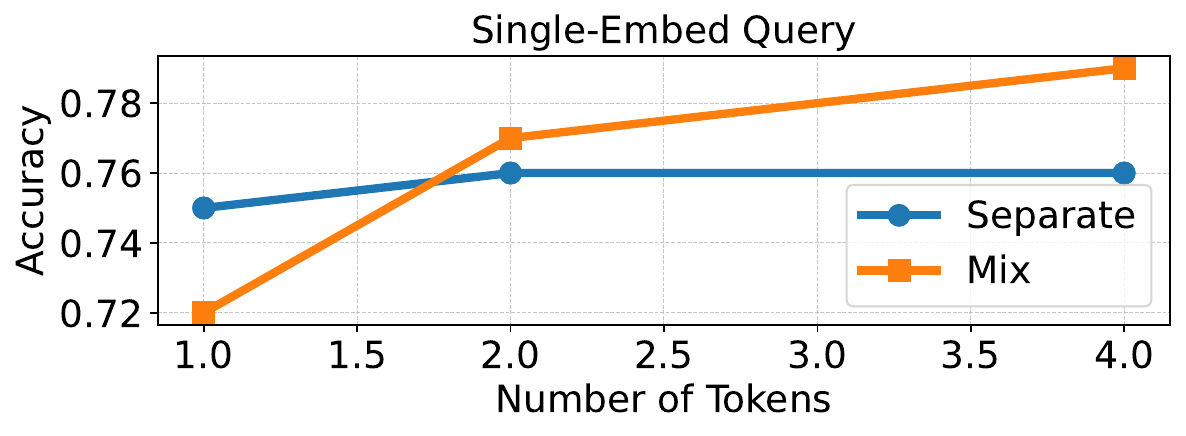}
    \vspace{-8pt}
  \end{subfigure}
    ~
  \begin{subfigure}[b]{0.32\linewidth}
    \centering
    \includegraphics[width=1\linewidth]{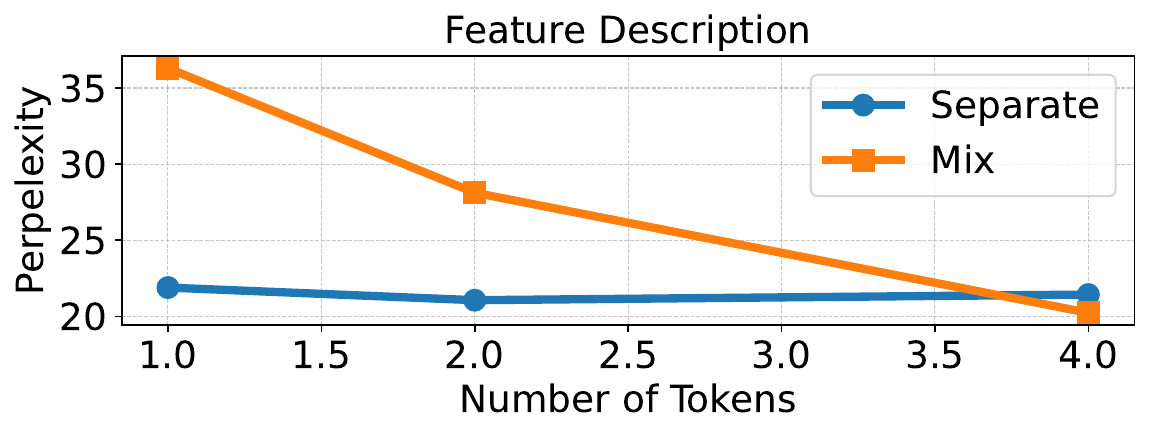}
    \vspace{-8pt}
  \end{subfigure}
  ~
  \begin{subfigure}[b]{0.32\linewidth}
    \centering
    \includegraphics[width=1\linewidth]{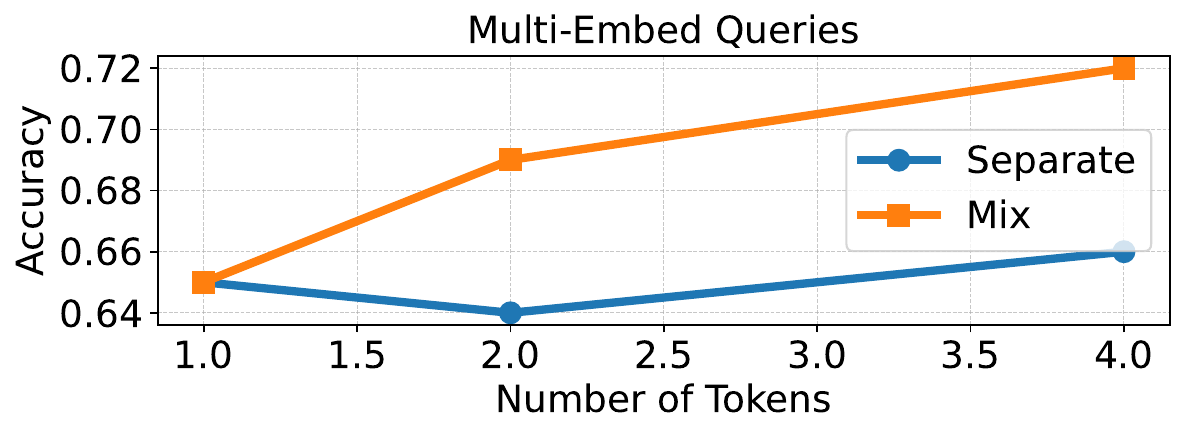}
    \vspace{-8pt}
  \end{subfigure}
\vspace{-10pt}
  \caption{Effect of soft token count $N$ under Separate and Mix training strategies.}
  \label{fig:qa_results_num_token}
\end{figure*}

\subsection{Experimental setup}
Unless otherwise specified, all experiments are conducted using gemma-3-4b-it as backbone. Our primary dataset consists of 7,000 unique samples, partitioned into a training set of 6,000 instances and a test set of 1,000. To prevent data leakage, there is no geographic overlap between the regions present in the training and test sets. A detailed experimental setup is shown in \Cref{app:setup}.

\textbf{Baseline} To quantify the specific advantage of our DFR-Gemma approach, we contrast it against several competitive baselines. A more direct pipeline visualization is shown in \Cref{fig:fig1}. Examples are shown in \Cref{app:baseline_data}
\begin{itemize}
\item \textbf{Zero-Context (Base Model Prior):} Evaluates the LLM’s intrinsic knowledge without external inputs. Performance reflects prior knowledge and biases (e.g., more coffee shops than hospitals) and serves as a baseline to measure gains from geospatial features.

\item \textbf{Unprocessed Raw Input:} Feeds raw feature names and values directly to the model. Inputs exceeding the context window are truncated, and numerical data is highly token-inefficient (e.g., numericals require 12–20 tokens due to digit-level tokenization).

\item \textbf{Raw Data Description:} Uses \texttt{gemma-3-4b-it} to summarize the top features into natural language under context constraints. While more readable, this approach remains limited by summarization quality, token inefficiency, and numerical errors from tokenization. Detailed processing pipeline and prompt are shown in \Cref{app:data_description}

\item \textbf{No LLM:} Trains task-specific models (e.g., MLP, LightGBM \cite{NIPS2017_6449f44a}) directly on PDFM embeddings, following \cite{agarwal2024general}. This isolates the benefit of LLM-based reasoning and quantifies the reasoning gain from feature–LLM alignment.
\item \textbf{Fragmented Pipeline:} The Fragmented Pipeline decomposes reasoning into multiple stages, where PDFM embeddings are first used for retrieval or processed by intermediate models, and the resulting outputs are then converted into text for LLM reasoning. Such pipelines require task-specific components (e.g., retrievers or predictors). For tasks that do not involve retrieval (e.g., single-embedding queries), these methods degenerate to the data description baseline. Therefore, we evaluate fragmented pipelines only on tasks that require multi-step retrieval and reasoning.
\vspace{-5pt}
\end{itemize}
\subsection{Results}
We evaluate the DFR framework on a diverse set of geospatial reasoning tasks. These represent high-utility, real-world queries. For example, identifying which coffee shops in a user’s region are currently less crowded. 

\begin{wrapfigure}{r}{0.45\textwidth}
\centering
\includegraphics[width=0.45\textwidth]{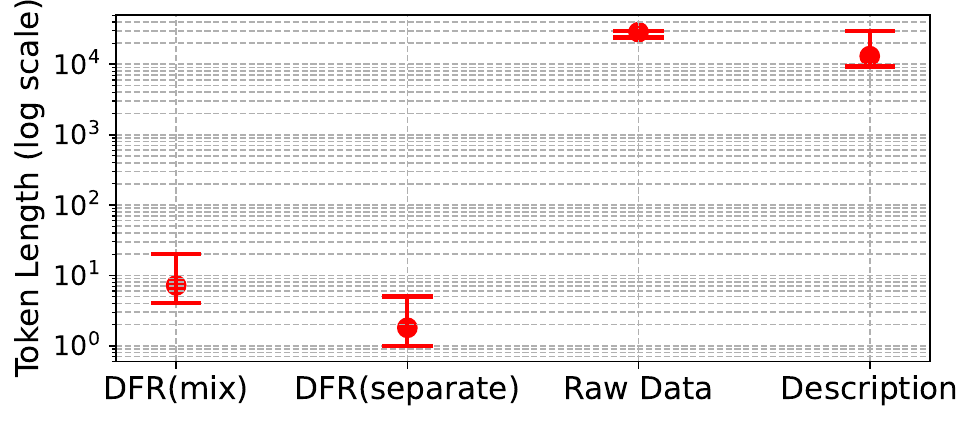}
\vspace{-10pt}
\caption{Token length distribution (min–max range with average marker).}
\label{fig:length}
\end{wrapfigure}
\textbf{Overall Performance:} The results across all tasks (Table~\ref{tab:qa_results}) show that DFR-Gemma consistently outperforms all baselines. While improvements over the Zero-Context baseline reflect gains from external data, adding data alone does not guarantee better performance. Naive approaches can degrade results (e.g., Unprocessed Raw Input drops from 67\% to 63\%), indicating that they introduce semantic noise. In contrast, DFR-Gemma bypasses textualization and operates directly in the latent space, enabling more effective reasoning. The No LLM baseline confirms that PDFM embeddings are highly informative; however, DFR-Gemma(Mix, N=4) surpasses it by up to 33\% on complex multi-embedding tasks, demonstrating a clear \emph{reasoning premium} from aligning features with LLMs. Compared to text-based baselines (Raw Data Description, Unprocessed Raw Input), DFR achieves consistent gains, especially in complex settings of multi-embedding queries. Notably, even with additional LLM cost, Raw Data Description fails to match DFR, highlighting the advantage of direct feature reasoning.

We further validate architectural generality by applying DFR to Qwen-2-4B, achieving consistent improvements across baselines. Finally, as shown in Table~\ref{fig:length}, DFR-Gemma significantly reduces input length compared to text-based methods, lowering computational cost while increasing information density.

\begin{wrapfigure}{r}{0.5\textwidth}
\centering
\includegraphics[width=0.5\textwidth]{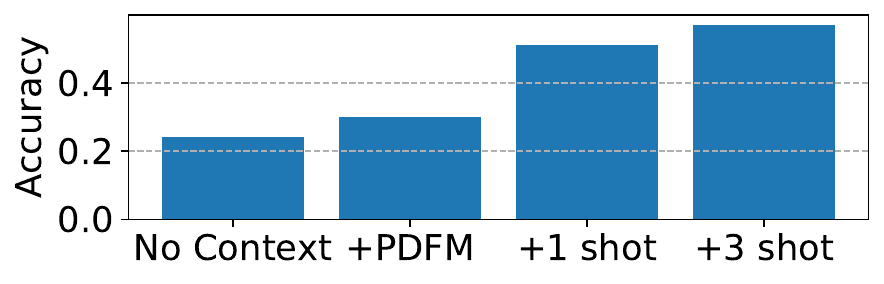}
\vspace{-10pt}
\caption{Absolute Value Estimation using 4-Option Multiple-Choice Evaluation.}
\label{fig:absolute}
\vspace{-5pt}
\end{wrapfigure}
\textbf{Joint Reasoning over Embedding and Text:} 
Beyond intrinsic reasoning, we evaluate DFR-Gemma’s ability to jointly reason over embeddings and external text. As shown in Table~\ref{fig:absolute}, the model initially struggles to infer absolute values (e.g., coffee shop counts) from PDFM embeddings, which primarily encode relative patterns. However, performance improves substantially from 0.30 to 0.57 with just three-shot textual examples. This indicates that the LLM can use in-context learning to calibrate embeddings with textual knowledge, enabling accurate reasoning over both representations. This flexibility enables seamless extension to additional modalities (e.g., images) and provides a practical mechanism for adapting to distribution shifts (shown later in "Generalizability to Distributional Shifts") via lightweight contextual calibration.

\textbf{Multi-Task Synergy and Token Capacity:}
To justify our design, we compare two training strategies: \emph{Separate}, which trains task-specific projectors, and \emph{Mix}, which uses a single projector across all tasks (our default setting). We further analyze the effect of projecting each PDFM embedding into $N$ tokens by evaluating how increasing $N$ impacts performance. As shown in \Cref{fig:qa_results_num_token}, the Separate strategy achieves strong performance with $N=1$, indicating that a single token suffices for individual tasks. In contrast, Mix benefits from increased token capacity: while $N=1$ is insufficient to capture diverse task requirements, expanding to $N=4$ significantly improves performance, surpassing task-specific projectors on complex settings (e.g., +6\% on multi-embedding queries). This suggests that multi-task training promotes more generalizable feature extraction, allowing tasks to share and reinforce underlying representations.
\begin{table*}[t]
\centering
\resizebox{1\textwidth}{!}{%
\begin{tabular}{l c c c c c } 
\toprule
\textbf{Method} & 
\textbf{Single-Embed Queries}& 
\textbf{Feature Description} & 
\textbf{Multi-Embed Queries} & \textbf{HellaSwag} &\textbf{GPQA Diamond}\\

\midrule
DFR-Gemma & 0.75 & 21.89  & 0.65  & 0.77$\star$ & 0.15$\star$\\
Projector+First layer & 0.78 & 20.71  & 0.63  & 0.66&0.09\\
Projector+Full LLM & 0.76  & 15.64  & 0.67  & 0.53 & 0.04\\
\bottomrule
\end{tabular}
}
\caption{Compare DFR-Gemma (only the projector is trainable) against two increasingly intensive training baselines. $\star$ are results from \cite{gemma_3n_2025}, which is the official report. HellaSwag~\cite{zellers2019hellaswag} and GPQA Diamond~\cite{rein2024gpqa} are benchmarks commonly used to evaluate reasoning capabilities in language models.}
\label{tab:qa_results_finetune}
\end{table*}

\textbf{Efficiency and Reasoning Preservation:} 
To evaluate the impact of different levels of model adaptation, we compare DFR-Gemma against two increasingly intensive training baselines: (1) Projector + First Layer, where only the MLP and the initial transformer block are optimized, and (2) Projector + Full LLM, which involves end-to-end supervised fine-tuning (SFT) of the entire architecture. As shown in Table~\ref{tab:qa_results_finetune}, our DFR approach, which keeps the entire LLM backbone frozen, achieves accuracy comparable to both unfrozen settings on Geo-reasoning tasks. While unfreezing the first layer or the full model allows for a more aggressive adaptation to PDFM features, it risks catastrophic forgetting of the LLM’s pre-trained linguistic and logical priors. For the unfrozen settings, the performance on reasoning task HellaSwag and GPQA Diamond clearly drop after finetuning on Geo-reasoning tasks, (77\% to 53\% and 15\% to 4\%). DFR-Gemma successfully bridges the modality gap while maintaining the model’s core reasoning capabilities. This makes DFR a more stable and parameter-efficient framework for specialized geospatial intelligence.

\begin{table*}[t]
\centering
\resizebox{0.8\textwidth}{!}{%
\begin{tabular}{ll c  c} 
\toprule
\textbf{Type} & \textbf{Method} & 
\textbf{Single-Embed}& 
\textbf{Multiple-Embed}\\

\midrule
\multirow{3}{*}{No Training} & Zero Context & 0.61 (-0.06)   & 0.14 (-0.07) \\
& Unprocessed Raw Input &  0.57 (-0.06)   & 0.13 (-0.05)  \\
& Raw Data Description &  0.66 (-0.04)   &  0.42 (-0.04)  \\
\midrule
\multirow{2}{*}{Ours} & Separate, N=1& 0.74 (-0.01)   & 0.62 (-0.03) \\
& Mix, N=4 &  0.79 (+0)  &  0.73 (+0.01) \\

\bottomrule
\end{tabular}
}
\caption{Evaluation of Model Stability across Linguistic Styles. Accuracy scores are shown for out-of-domain (OOD) writing styles, with the change from the original baseline performance noted in parentheses. DFR-Gemma maintains near-perfect consistency (marginal to zero delta), whereas text-based baselines experience significant drops due to sensitivity to syntactic and stylistic shifts.}
\label{tab:qa_results_ood_writing}
\end{table*}

\textbf{Robustness to Linguistic Variance:} To evaluate the model's resilience against out-of-domain phrasing, we subjected the queries to two distinct stylistic perturbations:
\begin{itemize}
    \item \textbf{Formal Academic Style:} Queries are restructured using low-frequency academic lexicon and complex syntax (e.g., \texttt{"In the municipality of Mountain View, does the density of coffee-oriented establishments exceed that of milk tea vendors?"}).
    \item \textbf{Noisy / Informal Internet Style} Rewrite the question in an informal internet style, allowing mild typos, abbreviations, and relaxed grammar, while keeping it understandable. \texttt{“in mountain view r there more coffee shops or milk tea shops lol?”}
\end{itemize}
As shown in \Cref{tab:qa_results_ood_writing}, DFR demonstrates significantly higher robustness to stylistic variations than all baselines. In contrast, baseline methods are highly sensitive to wording. For Zero-Context, small phrasing changes often trigger incorrect internal associations, leading the model to rely on spurious priors rather than the intended query logic. For Unprocessed Raw Input and Raw Data Description, stylistic shifts frequently cause attention drift: the model must parse text and numerical tokens with noisy information, and changes in phrasing can distract attention away from the relevant features.

DFR-Gemma remains stable because it reasons over embeddings rather than literal text, effectively decoupling spatial facts from linguistic form. The geospatial signal is encoded in a fixed set of soft tokens, allowing the model to focus on a consistent representation even when the query style varies. Moreover, the Mix (shared projector) configuration shows stronger robustness than Separate, as multi-task training encourages a more generalized alignment and reduces sensitivity to specific phrasing patterns.



\begin{figure*}
\centering
  \begin{subfigure}[b]{0.45\linewidth}
    \centering
    \includegraphics[width=1\linewidth]{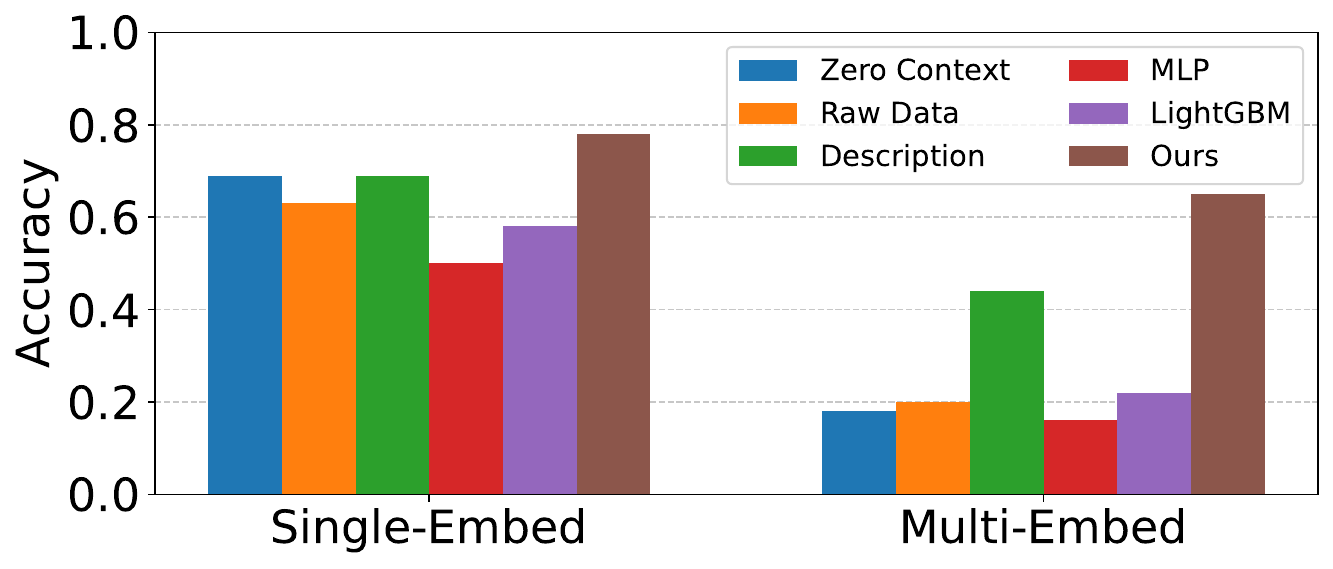}
    \vspace{-8pt}
  \end{subfigure}
  ~
  \begin{subfigure}[b]{0.45\linewidth}
    \centering
    \includegraphics[width=1\linewidth]{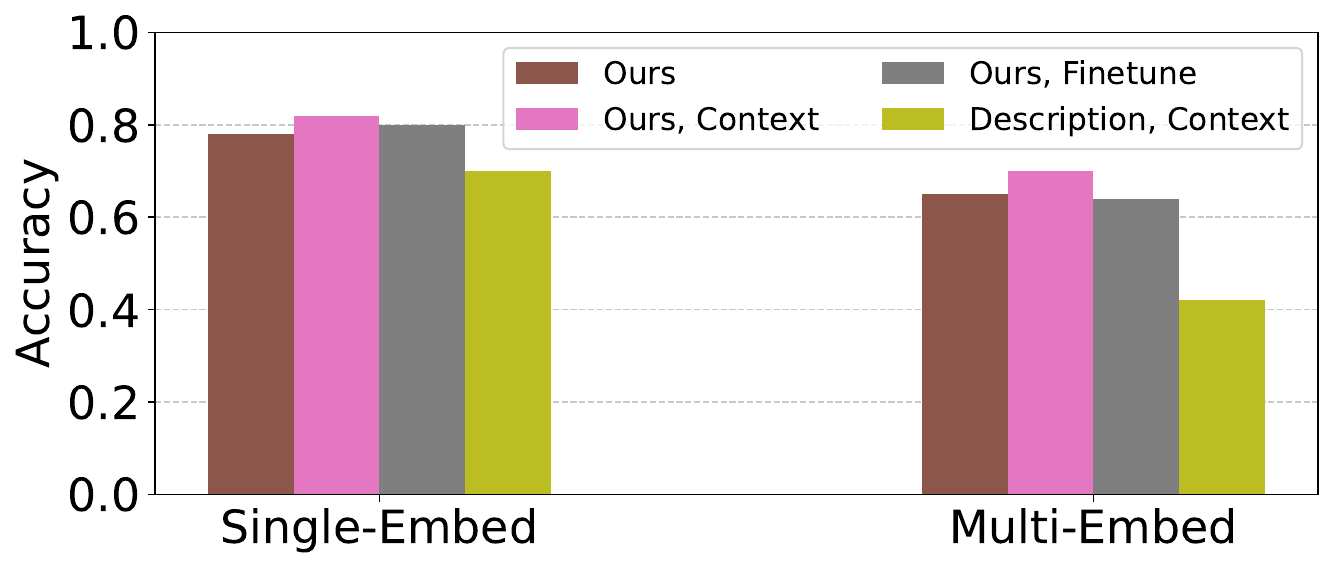}
    \vspace{-8pt}
  \end{subfigure}
  \vspace{-5pt}
  \caption{Model Robustness to Geographic Distribution Shifts. We evaluate the transferability of the DFR framework from postal-code-level training to county-level testing.}\vspace{-5pt}
  \label{fig:qa_results_ood_domain}
\end{figure*}

\textbf{Generalizability to Distributional Shifts:}
We evaluate DFR-Gemma under distributional shift by transferring from postal-code-level embeddings to coarser county-level embeddings, both encoded by the same PDFM. As shown in Table~\ref{fig:qa_results_ood_domain} (left), DFR-Gemma remains robust, while baselines degrade significantly, especially non-LLM models (MLP, LightGBM).

We further show that DFR-Gemma adapts efficiently to new distributions using lightweight strategies (Table~\ref{fig:qa_results_ood_domain}, right). For Contextual Adaptation, adding a few-shot textual context improves accuracy from 0.78 to 0.82 on single-embedding tasks without parameter updates, indicating effective in-context calibration. In contrast, Raw Data Description with few-shot examples, despite being the strongest baseline, fails to benefit due to long, verbose inputs that create an information bottleneck, which DFR’s compact embeddings avoid. Finally, the parameter-efficient projector enables Few-Shot Fine-Tuning on a small set of target-domain samples, further stabilizing performance. These results highlight DFR as a flexible, low-cost approach for adapting to real-world distribution shifts. 

\begin{wrapfigure}{l}{0.36\textwidth}
\centering
\vspace{-10pt}
\includegraphics[width=0.36\textwidth]{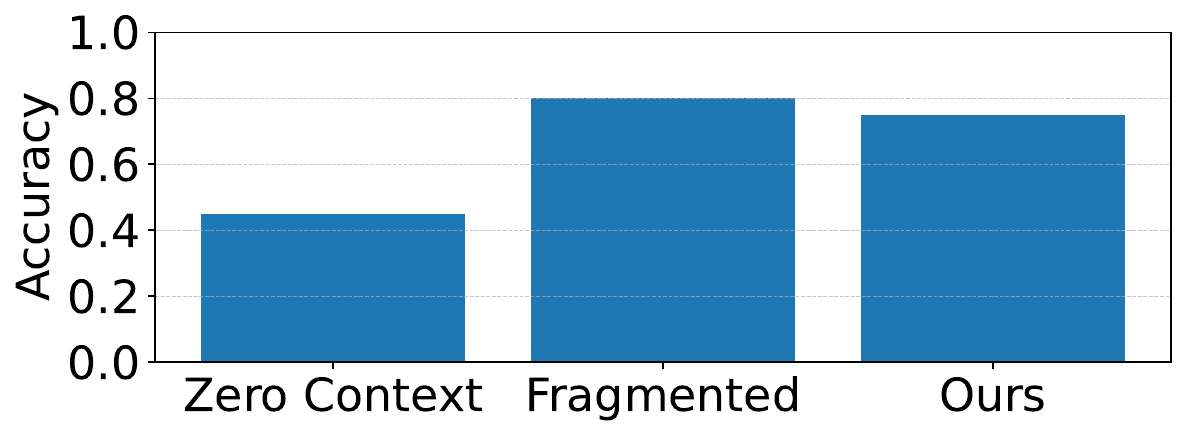}
\vspace{-20pt}

\caption{Multi-hop Reasoning.}
\vspace{-5pt}
\label{fig:multi-hop}
\end{wrapfigure}
\textbf{Multi-hop Reasoning and Comparison with Fragmented Pipelines:} To evaluate performance on complex, multi-step queries, we curated a specialized evaluation set requiring both spatial comparison and factual attribution. An example is:
\texttt{
Given the feature vector \texttt{<emb>}$e_0$\texttt{</emb>}, identify the most similar region in coffee shop distribution among \texttt{<emb>}$e_1$\texttt{</emb>}, 
\texttt{<emb>}$e_2$\texttt{</emb>}, and 
\texttt{<emb>}$e_3$\texttt{</emb>}, then determine whether its weather is hotter than the national average. Answer with "yes" or "no".
}
We build the fragmented pipeline using LightGBM-based retrievers. Since retrieval is feature- and objective-specific (e.g., most vs. least similar), we train separate models for each setting. Data of the retrieved region is then converted into a textual description and passed to the LLM for final reasoning. DFR-Gemma required no additional training or external modules, reasoning directly over the interleaved embeddings.

As shown in \Cref{fig:multi-hop}, no context reduces the LLM to near-random guessing. While the fragmented pipeline requires multiple specialized retrievers and staged processing, it only matches the performance of DFR-Gemma. In contrast, DFR-Gemma achieves comparable results within a single unified model, directly reasoning over embeddings without task-specific components or intermediate representations.

\section{Conclusion}

We introduced Direct Feature Reasoning-Gemma (DFR-Gemma), a framework that enables LLMs to reason directly over dense geospatial embeddings by aligning them with the model’s latent space. Our results show that direct embedding integration improves accuracy, token efficiency, and robustness compared to textualization and RAG-based pipelines. These findings highlight a shift from treating embeddings as auxiliary signals to using them as primary inputs for reasoning.

A key limitation of our approach is its reliance on high-quality geospatial foundation models (e.g., PDFM), as the effectiveness of reasoning depends on the richness of the underlying embeddings. Future work will extend this paradigm to additional modalities, such as temporal dynamics and satellite imagery, toward more general geospatial intelligence.














\bibliography{main}
\newpage
\appendix
\section{Detailed Experimental Setup} \label{app:setup}
To evaluate the generalizability of the Direct Feature Reasoning (DFR) framework across diverse architectural paradigms, we conduct experiments using two distinct state-of-the-art LLM families: Gemma-3-4B-IT ($d_{llm}=2048$) and Qwen3-4B-Instruct-2507 ($d_{llm}=4096$).

The DFR-Projector $\phi$ is implemented as a two-layer bottleneck Multi-Layer Perceptron (MLP) designed to align the continuous manifold of the PDFM with the discrete semantic space of the LLM. The architecture consists of an input dimension matched to the PDFM’s feature size ($d_e$) and an output dimension aligned with the LLM’s hidden state ($d_{llm}$). We utilize a hidden dimension $d_{mid}$ set to exactly half of the LLM's latent dimension ($d_{llm}/2$), employing a Gaussian Error Linear Unit (GELU) activation function for non-linear mapping. 

Unless otherwise specified, we employ a shared projector across all data modalities (a "mixed" configuration) and configure it to output a single "soft token" ($N=1$) per PDFM embedding to maintain maximum token efficiency. 

For all experiments, we adopt the Rotary Positional Embedding (RoPE) scheme, which serves as the native positional encoding mechanism for both the Gemma and Qwen backbones. To ensure structural consistency when interleaving aligned soft tokens with discrete linguistic tokens, we implement a sequence re-indexing protocol. This maintains relative positional integrity by re-computing the position\_ids for the entire mixed-modality sequence, allowing the LLM to correctly interpret the spatial relationship between interleaved latent features and text instructions. 

All training procedures are conducted on a cluster of eight NVIDIA A100 (80GB) GPUs on Google Cloud Platform.

\section{Data Collection and QA Generation} \label{app:data}
\subsection{QA Generation Pipeline}
To facilitate cross-modal learning, we constructed a comprehensive multi-task geospatial dataset pairing PDFM embeddings with verifiable question and ground-truth answer pairs. The generation pipeline consists of three primary stages:
\begin{enumerate}
    \item \textbf{Ground-Truth Extraction:} We extract a diverse set of raw features that the PDFM embeddings are hypothesized to encode, including environmental metrics (e.g., weather patterns), localized activity levels (busyness), and digital intent signals (search frequency) mapped to specific postal codes.
    \item \textbf{Synthetic QA Synthesis:
    }For every feature pair, we programmatically generate distinct question formats. Example: \texttt{"How does the [Feature] in [Postal Code] compare to the national average?" $\rightarrow$ "Higher".} 
    \item \textbf{Semantic Augmentation:} To prevent the model from overfitting to rigid templates, we rewrite these pairs into diverse, natural language variations. Original: \texttt{"How does the [Feature] in [Postal Code] compare to the national average?" $\rightarrow$ "Higher".} Augmented: \texttt{Is the [Feature] level in [Postal Code]
higher than the national average level?$\rightarrow$ Yes"}
\end{enumerate}

\subsection{Included tasks} \label{app:data_task}
The DFR-Gemma architecture is trained and evaluated using various geospatial reasoning tasks. These tasks are designed to teach the model to decode and reason over geospatial embeddings, bridging the gap between high-dimensional embeddings and natural language.
\begin{itemize}
    \item \textbf{Single-embedding, PDFM Internal Feature Query Test}: Single-embedding queries designed to test Gemma's understanding of the PDFM after alignment. The question must only concern internal features of  single PDFM content. The task should act as a direct verification step to confirm that Gemma is correctly processing and extracting information only from the aligned PDFM embedding. There are three types of queries:
    \begin{itemize}
        \item Compared to average: the value in a specific region is lower or higher than average.  Example: \texttt{"As shown in \texttt{<emb>}$e_0$\texttt{</emb>}, how does the [Feature] in postal code [Postal Code] compare to the national average?" $\rightarrow$ "Lower"}
        \item  Feature comparison: compare the feature values in a specific region. Example: \texttt{"As shown in \texttt{<emb>}$e_0$\texttt{</emb>}, in postal code [Postal Code] are there more coffee shops or more milk tea shops?"$\rightarrow$ "coffee shops"}
        \item  Ask absolute value: Example: \texttt{"As shown in \texttt{<emb>}$e_0$\texttt{</emb>}, how many [Feature] in postal code [Postal Code]?" $\rightarrow$ "21" }
    \end{itemize}
    \item \textbf{Single-embedding, PDFM Internal Feature Description}: This task evaluates the model's ability to synthesize PDFM embeddings into coherent, natural language summaries, similar to image captioning. Using structured prompts with length constraints, we generate ground-truth regional descriptions for both training labels and evaluation targets.
    \item \textbf{Multiple embedding, PDFM Internal Feature Query Test}: Test the model's ability to extract information and reason with multiple PDFM embeddings, e.g., from different regions for time. Input: Provide Gemma with multiple PDFM embedding alongside a structured instruction. Ask the model to reason based on multiple PDFM embeddings. 
    \begin{itemize}
        \item Find a similar region: find which region is most/least similar to target region. Example: \texttt{"Given the feature vector \texttt{<emb>}$e_0$\texttt{</emb>} for [Postal Code 0], which of the feature vectors \texttt{<emb>}$e_1$\texttt{</emb>} ([Postal Code 1]), \texttt{<emb>}$e_2$\texttt{</emb>} \\([Postal Code 2]), \texttt{<emb>}$e_3$\texttt{</emb>} ([Postal Code 3]), or \texttt{<emb>}$e_4$\texttt{</emb>} ([Postal Code 4]) is the most similar, where the features represent weather?" $\rightarrow$ "[Postal Code 4]" }
        \item Ask absolute value with other regions as the context. Example: \texttt{ "Given the PDFM embedding \texttt{<emb>}$e_0$\texttt{</emb>} for Zip Code [Postal Code 0] (which contains 10 coffee shops) and the PDFM embedding \texttt{<emb>}$e_1$\texttt{</emb>} for [Postal Code 1] (which contains 8 coffee shops), how many coffee shops are in the embedding \texttt{<emb>}$e_2$\texttt{</emb>} for Zip Code [Postal Code 2]?" $\rightarrow$ "12" }
        \item Compare feature value in different regions. Example: \texttt{"Given the PDFM embedding \texttt{<emb>}$e_0$\texttt{</emb>} for Zip Code [Postal Code 0] and the PDFM embedding \\\texttt{<emb>}$e_1$\texttt{</emb>} for Zip Code [Postal Code 1], which one contains more \\coffee shops?" $\rightarrow$ Zip Code [Postal Code 1].}
    \end{itemize}
\end{itemize}

\section{Baseline Context Example} \label{app:baseline_data}
\subsection{Unprocessed Raw Input Generation} \label{app:data_Raw}
\begin{lstlisting}[style=PromptStyle]
Input Variables: > - Region: [Postal Code 90210]

Feature A: [Population Density] | Value: [12,000]

Feature B: [Median Home Value] | Value: [$1.2M]
\end{lstlisting}

\subsection{Data description generation} \label{app:data_description}
To generate natural language baselines, we utilize Gemma-3-4B-IT to synthesize raw geospatial signals into structured summaries. We restrict this synthesis to the top 20 most significant features per region. This threshold is strategically chosen to balance information coverage with the architectural constraints of the LLM; expanding the description beyond 20 features significantly inflates the prompt length, frequently exceeding the effective context window when multiple regions are compared. Furthermore, our empirical observations indicate that including a higher cardinality of raw features often degrades reasoning performance, as the model struggles with the inherent noise of long-form numerical text.

While these descriptions improve human readability, this approach remains fundamentally bottlenecked by three factors: (1) the subjective quality of the summarizer, (2) the extreme token-inefficiency of translating dense vectors into prose, and (3) persistent numerical reasoning failures caused by tokenizer fragmentation, where multi-digit values are split into incoherent sub-tokens that obscure the underlying quantitative relationships.

The following is the generation prompt. The prompt is written and improved with Google Gemini.

System Prompt:
\begin{lstlisting}[style=PromptStyle]
You are a Geospatial Data Analyst. Your task is to transform raw numerical feature vectors into a concise, high-fidelity natural language summary. Focus on precision and preserve the exact numerical scale of each feature.
\end{lstlisting}

User Prompt:
\begin{lstlisting}[style=PromptStyle]
Task: Summarize the following top 20 geospatial features for [Region Name/ID].

Raw Data (Feature: Value):
[Insert List of 20 Features here, e.g., Population Density: 1240.5; Median Income: 65000; NDVI: 0.12...]

Instructions:

Start with a brief one-sentence overview of the region's profile.

Group the 20 features into 3 logical categories (e.g., Demographics, Environment, Infrastructure).

List each feature and its exact value. Do not round numbers or omit units.

Avoid flowery language; maintain a technical and objective tone.

Ensure the total length does not exceed 150 words to maintain context efficiency.

Summary:
\end{lstlisting}

Generated examples:
\begin{lstlisting}[style=PromptStyle]
Description for 0810: ## Senior Search Data Analyst Report - Location ID: 0810 (2025-10) **Executive Summary:** The search volume for Location ID 0810 in October 2025 is substantial, with a high concentration of searches relating to general information, entertainment, and consumer goods. The primary intent appears to be a blend of informational queries (definition, time, news) and transactional searches (price, sales, purchase-related terms), indicating a diverse user base seeking both knowledge and potential purchases within this location. **Top Themes:** 1. **Consumer Goods & Entertainment:** This is the most dominant theme, encompassing categories like "Car," "Phone," "TV," "Game," "Restaurant," "Movie theater," "Music," "Clothing," and "Food." This suggests a significant interest in purchasing and consuming various goods and services. 2. **Information & General Knowledge:** A considerable portion of searches revolve around general information, including "Definition," "Weather," "Time," "News," "Meaning," "Google," and "Map." This indicates users are actively seeking information about their surroundings and the world around them. 3. **Health & Well-being:** Categories like "Health," "Medicine," "Massage," and "Sleep" represent a notable segment of searches, pointing towards user interest in personal well-being and related services. **Anomalies:** Several search terms stand out as relatively frequent and potentially niche: * **"ALDI" and "Aldi":** The repeated appearance of "ALDI" suggests a strong interest in this specific grocery retailer within the location. * **"Japanese":**  Combined with other terms like "Travel," "Hotel," and "Restaurant," this indicates a potential interest in Japanese culture or travel within or to the location. * **"Alcohol":** While not explicitly listed, the presence of "Coffee" and "Alcohol" in the top 200 suggests a possible interest in social gatherings and beverage options. * **"Horse":** This is an unusual term within the top 200 and could represent local interests in equestrian activities or unique local attractions. **Actionable Insight:** Based on this data, a manager should recognize that users in Location ID 0810 exhibit a diverse range of interests, with a strong leaning towards consumer goods, entertainment, and general information seeking. The frequent searches for "ALDI" highlight a significant local interest in this specific retailer, while the broader informational queries suggest a user base actively engaged with their surroundings. This indicates a strong potential for targeted marketing campaigns focusing on local businesses and relevant information for this area. Further analysis could explore the relationship between these themes and specific demographics within Location ID 0810 to refine marketing strategies.
\end{lstlisting}

\section{Detailed PDFM explanation}\label{app:pdfm}
Here we show the framework of PDFM with example of input data.
\begin{figure}
\centering
\includegraphics[width=0.4\textwidth]{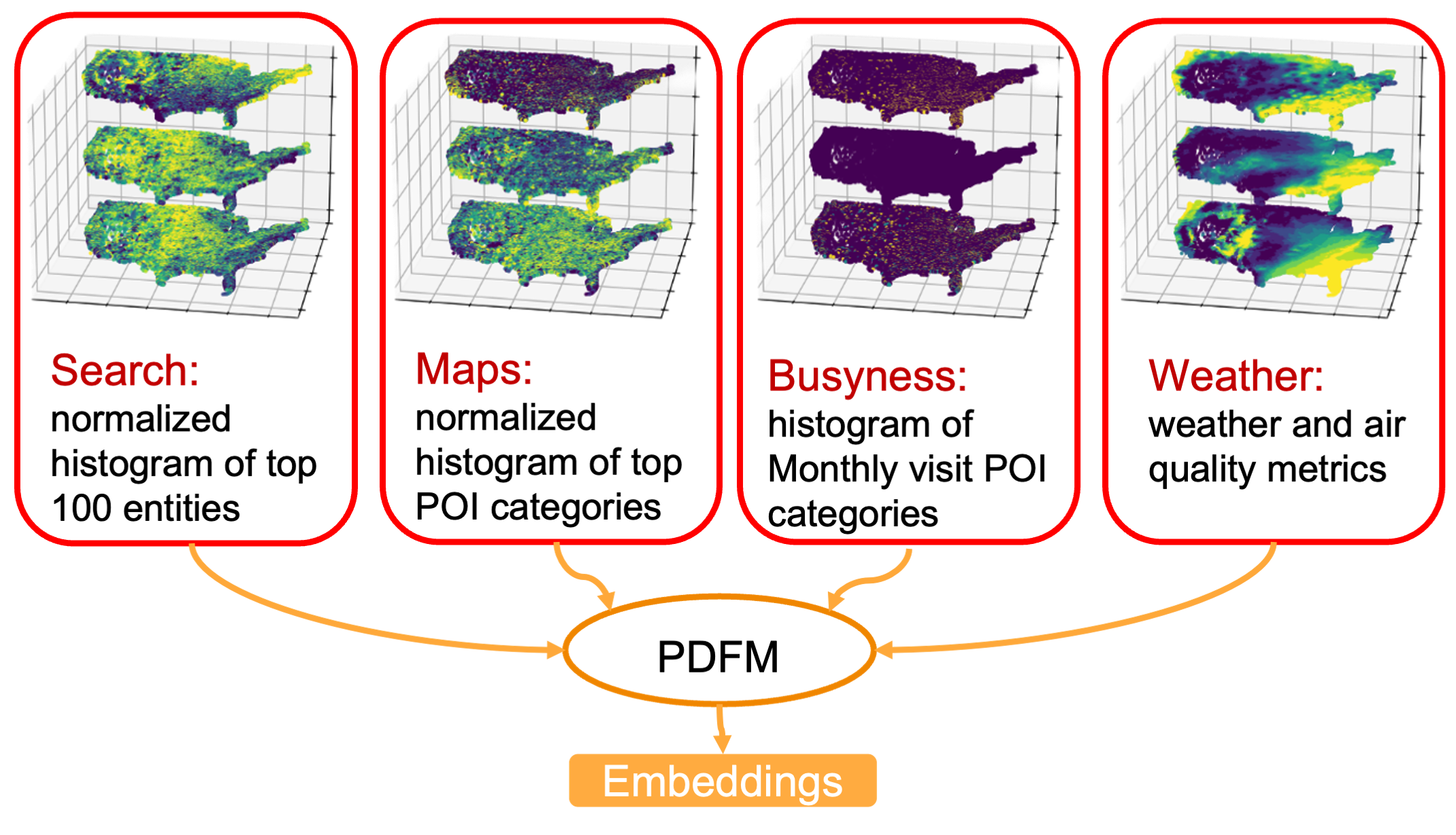}
\caption{\small{PDFM framework.}}
\label{fig:pdfm}
\end{figure}

\section{Implementation and Inference Pipeline}\label{app:pseudo_code}
The DFR inference pipeline is designed to be a "plug-and-play" extension to standard transformer-based generation. We bypass the discrete tokenizer and inject high-dimensional spatial semantics directly into the model's first layer.  


\begin{lstlisting}[style=PythonStyle]
def inference_pipeline(query, raw_pdfm, indices, model, tokenizer, projector):
    """
    Args:
        query: Natural language instruction string.
        raw_pdfm: High-dimensional PDFM feature vectors [Batch, K, d_e]. 
                  (K is the number of regions/points).
        indices: Placement indices for embedding injection.
        model: Frozen LLM Backbone (e.g., Gemma-3).
        projector: Trainable Cross-Modal Projector with terminal expansion.
    """
    # 1. Text Tokenization & LLM Embedding Layer lookup
    # Convert text to discrete IDs and then to continuous vectors.
    token_ids = tokenizer(query, return_tensors="pt").input_ids
    text_embeds = model.get_input_embeddings(token_ids) # Shape: [Batch, L, d_L]
    
    # 2. Cross-Modal Alignment (Projection)
    # Project each PDFM embedding into a sequence of N soft tokens.
    # Z_i = Reshape(MLP(e_i), (N, d_L))
    z_aligned = projector(raw_pdfm) # Shape: [Batch, K * N, d_L]
    
    # 3. Sequence Interleaving & Positional Re-indexing
    # Combine text embeddings and "soft" geospatial tokens into a unified sequence.
    # x_full: [Batch, L + (K * N), d_L]
    # pos_ids are recomputed to ensure structural consistency.
    x_full, pos_ids = interleave_and_reindex(text_embeds, z_aligned, indices)
    
    # 4. Intrinsic Reasoning and Generation
    # We pass 'inputs_embeds' to bypass the discrete tokenizer bottleneck.
    # Updated 'position_ids' ensure the LLM correctly interprets the interleaved sequence.
    output = model.generate(
        inputs_embeds=x_full, 
        position_ids=pos_ids,
        max_new_tokens=128
    )
    
    # Decode the generated tokens back into natural language.
    return tokenizer.decode(output)
\end{lstlisting}

    
    
    
    
\section{Hybrid Contrastive Learning (Alignment Tuning)}\label{app:Contrastive}

To evaluate whether explicit manifold structuring improves the reasoning capabilities of the \textbf{DFR-Projector}, we conduct an ablation study comparing our standard end-to-end training against a \textit{Hybrid Contrastive Learning} (Alignment Tuning) objective. While the default configuration optimizes the projector solely via the LLM's language modeling loss, the hybrid approach introduces a Supervised Contrastive (\textit{SupCon}) loss \cite{khosla2020supervised} to regularize the latent space.

We define the total objective function $\mathcal{L}_{\text{total}}$ as a weighted combination of the cross-entropy loss $\mathcal{L}_{\text{CE}}$ and the alignment loss $\mathcal{L}_{\text{align}}$:
\begin{equation}
    \mathcal{L}_{\text{total}} = \mathcal{L}_{\text{CE}} + \lambda \mathcal{L}_{\text{align}}
\end{equation}
where $\lambda$ is a hyperparameter (set to $0.1$ in our experiments). We employ the Supervised Contrastive (SupCon) loss \cite{khosla2020supervised} as $\mathcal{L}_{\text{align}}$. Given a batch of $N$ regional embeddings, let $i \in \{1 \dots N\}$ be the index of an anchor sample, and $P(i)$ be the set of indices of all samples in the batch sharing the same geospatial profile as $i$. The loss is defined as:

\begin{equation}
    \mathcal{L}_{\text{SupCon}} = \sum_{i \in I} \frac{-1}{|P(i)|} \sum_{p \in P(i)} \log \frac{\exp(z_i \cdot z_p / \tau)}{\sum_{a \in A(i)} \exp(z_i \cdot z_a / \tau)}
\end{equation}

where $z_i = \phi(e_i)$ represents the aligned soft token and $\tau$ denotes the temperature hyperparameter. We set $\tau=0.07$ in our paper. This objective encourages the projector to cluster regions with similar underlying characteristics in the LLM's latent space, thereby facilitating more robust downstream reasoning.

We conduct the experiments with Separate, N=1 on single-embedding query task. Shown in our results, the performnace decrease from 0.79 to 0.70 after adding $\mathcal{L}_{\text{CE}}$. The reukts are shown in \Cref{tab:con}.
\begin{table*}
\centering
\resizebox{0.3\textwidth}{!}{%
\begin{tabular}{l c  } 
\toprule
\textbf{Method} & 
\textbf{Accuracy}\\
\midrule
$\mathcal{L}_{\text{CE}}$ & 0.79\\
$\mathcal{L}_{\text{total}}$ & 0.70\\
\bottomrule
\end{tabular}}
\vspace{-5pt}
\caption{\small{Hybrid Contrastive Learning Performance.}}
\label{tab:con}
\vspace{-5pt}
\end{table*}

\section{Multi-Task Capacity via Selective Attention} \label{app:MultiTask}
A core design choice in our methodology is the projection of a single PDFM embedding into $N > 1$ soft tokens. While a $1:1$ mapping (where $N=1$) might seem more token-efficient, we argue that it creates a critical representation bottleneck that hinders multi-task performance.

\textbf{Information Density and the Bottleneck:} PDFM embeddings are highly compressed, multi-modal representations containing disparate signals such as commercial POI density, mobility-driven "busyness" levels, and environmental metrics. Compressing this multi-faceted information into a single LLM hidden state forces the projector to produce an "average" representation, which often loses the granular variance needed for specific reasoning tasks.\textbf{Task-Specific Selective Attention:} Our framework utilizes a single universal projector across a diverse multi-task benchmark. Different queries require the model to attend to different "latent dimensions" of the same geographic region. For example:\begin{itemize}[nosep,leftmargin=*]\item \textit{Socio-economic queries} require the model to decode commercial infrastructure signals.\item \textit{Dynamic mobility queries} require focusing on temporal busyness patterns.\end{itemize}By mapping each embedding to $N$ tokens, we provide the LLM with a larger "semantic surface area." This allows the Transformer's self-attention mechanism to perform \textit{selective extraction}: the model can learn to attend to tokens $z_{i,1 \dots n}$ for infrastructure-related questions while shifting its attention weights to $z_{i, n \dots N}$ for mobility-related questions. This multi-token representation ensures that the LLM can "query" the embedding for task-relevant information, effectively bypassing the capacity limits of a single-token bottleneck and enabling robust performance across our entire multi-task suite.
\end{document}